\documentclass{article}
\usepackage[utf8]{inputenc}

\usepackage{authblk}%https://preview.overleaf.com/public/mzkbzvzbnnhr/images/e125d7a4b943a77d9c348b390f7731b157bc46ad.jpeg
\usepackage{srcltx}
\usepackage[psamsfonts]{amsfonts}
\usepackage{amsmath}
\usepackage{amssymb}
\usepackage{graphics}
\usepackage{psfrag}
\usepackage{epsfig}
\usepackage{array}%https://preview.overleaf.com/public/mzkbzvzbnnhr/images/d03fd99d6fcd319a1965666c047c4193eab865c9.jpeg
\usepackage{algorithm}
\usepackage{algpseudocode}
\usepackage[psamsfonts]{amsfonts}
\usepackage{makeidx}  % allows for indexgeneration
\usepackage{bbm}
\usepackage{amsfonts}
\usepackage{hhline}
\usepackage{eufrak}
\usepackage{yfonts}
\usepackage{pifont}
\usepackage{footnote}
\usepackage{pdfpages}
\usepackage{graphicx}
\usepackage{balance}
\usepackage{color}
\usepackage[square, comma, sort&compress, numbers]{natbib}
\usepackage{dsfont}
\usepackage{caption}
\usepackage{subcaption}
\usepackage{sidecap}
\usepackage[splitrule,hang,marginal]{footmisc}
\usepackage{enumitem}
\usepackage{bm}

\title{final signal}
\author{sahar.tavakoli.85 }
\date{October 2020}

\begin{document}
 \pagenumbering{arabic}
\title{Signal Classification using Weighted Orthogonal Regression Method}
\author{ Sahar Tavakoli}

\maketitle

\begin{abstract}
%Most of the data mining problems could be categorized as a classification problem.
In this paper, a new classifier based on the intrinsic properties of the data is proposed. Classification is an essential task in data mining-based applications. The classification problem will be challenging when the size of training set is not sufficient compare to the dimension of the problem. This paper proposes a new classification method  which exploits intrinsic structure of each class through the corresponding eigen components. Each component contributes to the learned span of each class by a specific weight. The weight is determined by the associated eigenvalue. This approach results in a reliable learning robust to the case of facing a classification problem with limited training data. The proposed method involves the obtained Eigenvectors by SVD of data from each class to select the bases for each subspace. Moreover, it considers an efficient weighting for the decision making criterion to discriminate two classes. In addition to high performance  on artificial data, this method has increased the best result of an international competition.    
\end{abstract}
\vspace{-3mm}

\section{Introduction}
Each training datum in a supervised learning framework, consists of a feature vector $\boldsymbol{x}$ and its corresponding label $z$. The classification problem can be stated as generating a rule (classifier) $h$, so that $h(\boldsymbol{x})$ can estimate the class for any $\boldsymbol{x}$. It can be cast as follows
\begin{equation}
 \hat{z}=h(\boldsymbol{x}).  
\end{equation}

Where, the estimated labels are as close as possible to those of training data. The class label of test data are unknown; however, the learned classifier is able to estimate their labels. Design of a classifier implies solving the following optimization problem,
$
\hat{h}=\underset{h}{\text{argmin}}\sum_{i=1}^L d(z_i,h(\boldsymbol{x}_i)).
$
Where, $d$ is a distance function, e.g., Euclidean distance and $L$ is the number of train data. This problem can be interpreted as a regression problem. Regression-based classification partitions the sample data based on the given class labels. In order to learn reliable regions for each class we need to have large number of training data which is equivalent to the maximum likelihood classification which is an optimal method \cite{maximumlikelihoodbehinas}. This method is conditioned to have enough training samples for estimating the Probability Density Functions (PDFs) of the classes. However, there is a limitation in the number of training data in most problems. This limitation seems more critical in high dimensional problems. There is an exponential relationship between the dimension of a classification problem, and the number of needed training data to estimate the PDF \cite{rabetetrainingvabod}. Hence, estimating the PDF and applying the maximum likelihood classifier is impossible in most high dimensional problems. In such a case where the dimension of data is more than the dimension of unknown variables,  exploiting over-determined linear equation systems can be helpful. In this paper, first, solving of mentioned equation systems with sparseness constraints will be introduced, and second, its application in classification will be described in detail.
A linear equation system is called over-determined when the number of equations are more than the number of variables. A matrix form for this type of linear equation system is given below, 

\begin{equation}
\boldsymbol{x}=\boldsymbol{As},\qquad n>m.
\end{equation}
Where, $\boldsymbol{A}\in \mathbb{R}^{m\times n}$. This system of equations does not provide a unique solution, however, least square solution can be solved uniquely as follows,
$$
\boldsymbol{s}=\underset{\boldsymbol{s}}{\text{argmin}}\|\boldsymbol{x}-\boldsymbol{As}\|=(\boldsymbol{A}^T\boldsymbol{A})^{-1}\boldsymbol{A}^T\boldsymbol{x}
$$

Due to sensitivity of the above least square solution  to the condition number of $\boldsymbol{A}$, it is crucial to have a well-conditioned $\boldsymbol{A}$. While, in most machine learning application the training data are highly correlated, thus, $\boldsymbol{A}$ might be ill-conditioned. In order to stabilize the solution, the Ridge regression can be utilized \cite{hoerl1970ridge}. 
\begin{equation}
\label{ridge}
\min_{\bm{s}} \{{\lVert \bm{x}-\bm{As} \rVert}^2_2+\lambda{\lVert \bm{s} \rVert}_2^2 \}
\end{equation}
 $\lambda $ is a coefficient which determines the trade off between regularization term and regression error. The Ridge regression solution can be written as the following closed form,
\begin{equation}
\label{ridgesol}
\bm{s}=(\bm{A}^T\bm{A}+\lambda\bm{I})^{-1}\bm{A}^T\bm{x}
\end{equation}

The Ridge regression involves all of the training data in order to partition the realm of each data class. In many machine learning scenarios it is not desired to engage all of the training data. The first reason is avoiding over-learning, and second reason is exploiting the most relevant training data for regression \cite{joachims2006training,sah1,sah2,sah3,sah4}. This fact encourages us to consider sparsity in the regression problem. Inspired by sparsity the SVM classifier is suggested which contributes in data mining literature significantly. Moreover, the LASSO regression is one of the most well-known types of linear regression which is based on sparsity \cite{lasso}. LASSO minimizes the usual sum of squared errors, with a constraint on the sum of the absolute values of the coefficients. Mathematically speaking, Equation \eqref{lasso} shows the LASSO problem.
\begin{equation}
\label{lasso}
\min_{\bm{s}} \{{\lVert \bm{x}-\bm{As} \rVert}^2_2+{\lVert \bm{s} \rVert}_1 \}
\end{equation} 

Inspired by the two mentioned types of regression, Elastic-net type is introduced in form of the following equation .
\begin{equation}
\label{elastic}
\min_{\bm{s}} \{{\lVert \bm{x}-\bm{As} \rVert}^2_2+\lambda_1{\lVert \bm{s} \rVert}_1 +\lambda_2{\lVert \bm{s} \rVert}_2 \}.
\end{equation}
This regression promotes sparsity, however, the correlated training data effect the solution with a same sparsity pattern  .
\section{Related works}
Considering a two classes classification problem, one can arrange training data of a class as columns, and set a matrix that determines first class' subspace. In figure 1, $\bm{A}_1$ is a dictionary for first class, and its columns as basis vectors construct the first subspace. Similarly, $\bm{A}_2$ could determine the second subspace. 
\begin{figure}[H]
\label{figclass}
\includegraphics[width=.5\textwidth]{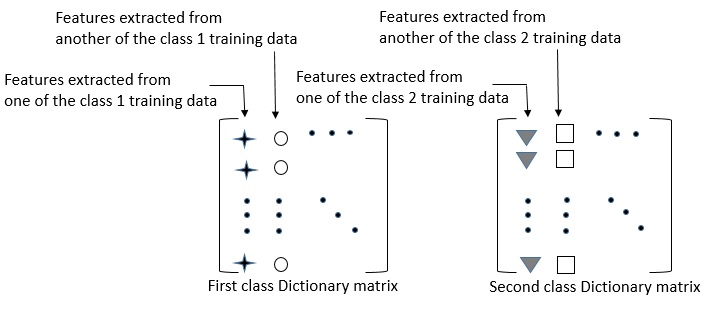}
\caption{creating Dictionary matrices for two classes classification problem}
\end{figure}
 Supposing that each test data which belongs to first subspace could be tagged by first class, and each test data which belongs to second subspace could be tagged by second class, one can easily make a decision for mentioned two classes classification\cite{subspaceclassification}. The decision strategy is shown in equation \eqref{decision}.
 
\begin{equation}
\label{decision}
  \begin{cases}
    \bm{H}_1:\bm{x}\in span(\bm{A}_1)\\
    \bm{H}_2:\bm{x}\in span(\bm{A}_2)
  \end{cases}
\end{equation}
 
 From now, the decision of the class for a test data will be equal to finding the subspace that the test data is belong to. The regression problem, and finding the coefficient vector for each subspace corresponding to the test data will be useful in this stage\cite{sep1}.  
 As it is known, the answer for regression problem in equation no. is the equation \ref{finds}.  
 \begin{equation}
 \label{finds}
\min_{\bm{s}} \{{\lVert \bm{x}-\bm{As} \rVert}^2_2\}
 \end{equation}
 \begin{equation*}
\bm{s}=(\bm{A}^T\bm{A})^{-1}\bm{A}^T\bm{x} 
 \end{equation*}
By  inserting sparseness constraint in the stage of solving the regression problem, and solving mentioned types of regression problem instead of a simple one name of the classifier will change to  LASSO, Ridge, or Elastic-net classifier .
The s vector which is a coefficient vector can be used to make a decision that a test data vector belongs to which class .Two different methods could be used to achieve a final decision\cite{sep2}. 
\subsection{Nearest Subspace Classifier}
First method is based on considering each subspace separately, and solving two regression problem, one for each subspace\cite{sep3}, and decide base on that two vectors according to equation \eqref{nearestdec}.
\begin{equation}
\label{nearestdec}
  \begin{cases}
    \bm{s}_1= (\bm{A}_1^T\bm{A}_1)^{-1}\bm{A}_1^T\bm{x}\\
    \bm{s}_2= (\bm{A_2}^T\bm{A_2})^{-1}\bm{A}_2^T\bm{x}
  \end{cases}
\end{equation}
\begin{equation}
\begin{cases}
{\lVert \bm{s}_1 \rVert}_\alpha > {\lVert \bm{s}_2 \rVert}_\alpha \Rightarrow Class1 \\
{\lVert \bm{s}_1 \rVert}_\alpha < {\lVert \bm{s}_2 \rVert}_\alpha \Rightarrow Class2
\end{cases}
\end{equation}
 In above relations, $\alpha$ could be any arbitrary number. 
One can also decides based on the reconstruction vectors. Each subspace that could rebuild the entrance vector would be the decided class for that vector.
Mentioned strategy could be easily understood from following equations. $\bm{x}_1$ and $\bm{x}_2$ are reconstructed vectors of entrance vector using respectively first and second subspace. 
\begin{equation}
  \begin{cases}
    \bm{x}_1= {\bm{A}_1}(\bm{A}_1^T\bm{A}_1)^{-1}\bm{A}_1^T\bm{x}\\
    \bm{x}_2= {\bm{A}_2}(\bm{A_2}^T\bm{A_2})^{-1}\bm{A}_2^T\bm{x}
  \end{cases}
\end{equation}
Comparing $\bm{x}_1$ and $\bm{x}_2$ to entrance vector, classifier can determine the winner which is the classes that has reconstructed the entrance vector with less difference. Therefore, the decision making strategy will be as follow:
\begin{equation}
\begin{cases}
{\lVert \bm{x} - \bm{x}_1 \rVert}_2 < {\lVert \bm{x} - \bm{x}_2 \rVert}_2 \Rightarrow Class1 \\
{\lVert \bm{x} - \bm{x}_1 \rVert}_2 > {\lVert \bm{x} - \bm{x}_2 \rVert}_2 \Rightarrow Class2
\end{cases}
\end{equation}
The nearest subspace classifier has the same structure, and consider each space separately to calculate the coefficients.    
\subsection{Union of Subspace Classifier}
 Unlike the nearest subspace classifier, the second method puts two dictionaries together, and makes a combined dictionary to solve one regression problem as indicated in equation \ref{union}\cite{union}.
 \begin{equation}
 \label{union}
 \bm{A} = [\bm{A}_1 \bm{A}_2]
 \end{equation}
  Equation \eqref{finds} indicates the mentioned regression problem. The result of this regression will be a combined coefficient vector as indicated in equation \eqref{s}. The $\bm{s}$ vector which is the solution of equation \eqref{finds} is combination of coefficients corresponding to each class as indicated in equation  \ref{s}.
  \begin{equation}
  \label{s}
  \bm{s} = [\bm{s}_1 \bm{s}_2]
  \end{equation} 
  Assuming that the number of testing data of first class is $n_1$, first coefficient until $n_1^{th}$ one will be corresponding to first class, and other ones will be corresponding to second class.

There for, instead of two regression problem, one problem with higher dimension could reach us to the coefficient vector of each class. From here with knowing the coefficient of each class, the decision strategy will be the same as before. This method has been named union subspace classifier. This approach, due to the processing on all training data, can give better results. 
Superiority of this method to the nearest subspace classification method is shown in the specific application of face recognition .
As mentioned, the real challenge in regression based classifier is to find the appropriate dictionary . Using all the data in the dictionary leads to over learning problem\cite{overlearning}, so choosing useful, and low number of bases would be our goal.

\section{Proposed Method}
Although applying the sparseness constraint in regression problem is useful in performance of classification, it will make the solving very difficult . 
The WORM method presented in this study is based on solving a regression problem without sparseness constraint in first stage, to increase computing convenience, and speed, and applying the sparseness in the second stage that is the decision phase to increase the classification quality.
This classification method has reached the desired result in testing on artificial data with Gaussian, salt and pepper, and multiplicative noise as well as experimental data. 
Following block diagram has described the general method of classification. 
\begin{figure}[h]
\includegraphics[width=.5\textwidth]{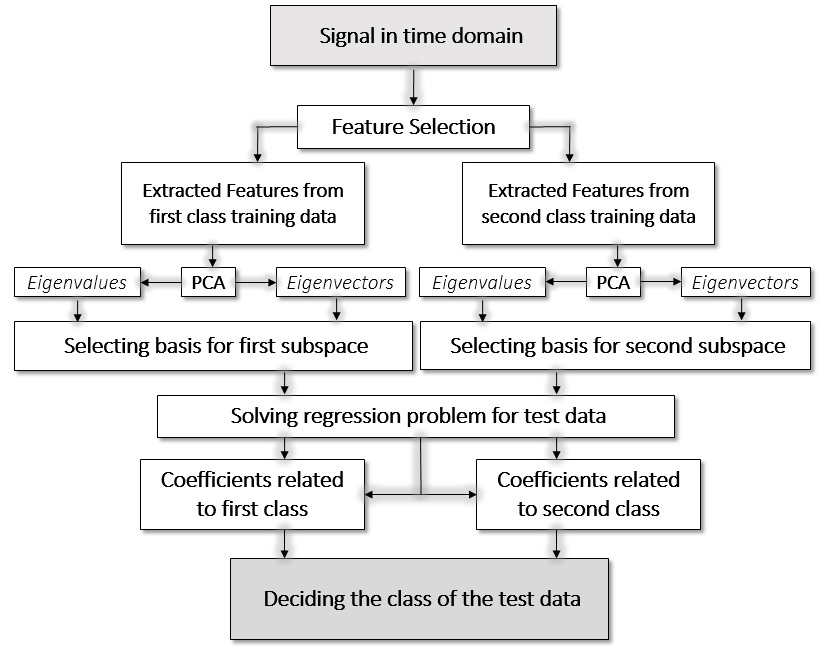}
\caption{Weighted Orthogonal Regression Method algorithm}
\end{figure}
First, the feature vector for each training data is extracted. As the problem is a regression type, growth of feature dimension will not be a negative point. Now, the appropriate basis for subspace of each class should be selected. This goal is achievable by PCA\cite{ az1, az2}. Although PCA has been used in several papers before, in all those articles, the aim of using PCA was feature reduction. However, PCA in this method is used to find the basis for each subspace which is one of the advantages of WORM.
As mentioned, to find an appropriate basis for each class is so important since using all the available data leads to over learning and reduction of classification quality.
By using PCA, not only proper basis which are the eigenvectors of the dictionary matrix is found, but also the number of required basis can be estimated based on the eigenvalues. 
The below chart shows the eigenvalues of dictionary matrix versus their number. The number of eigenvalues that provide the acceptable percentage of total energy indicates the optimal number of needed vectors as basis of the subspace.
\begin{figure}[h]
\includegraphics[width=.5\textwidth]{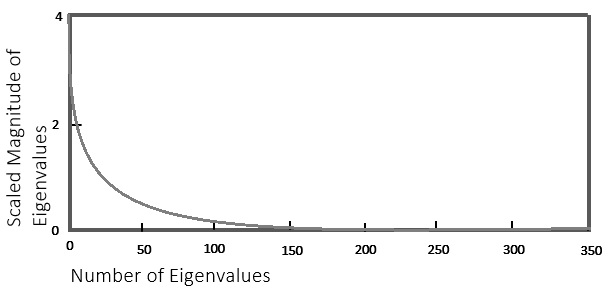}
\caption{The eigenvalues of dictionary matrix versus their number}
\end{figure}
After choosing proper basis for each class, those basis vectors are place together to form a dictionary matrix.
\begin{equation}
\bm{D} = [\bm{D}_1 \bm{D}_2]
\end{equation}
After finding the dictionary matrix, one regression problem will be solved on the overall dictionary matrix, and entrance test data that its class is unknown. 
In equation \eqref{regd} $\bm{y}$ is the mentioned test data, and the aim of the regression problem is to find the optimal coefficient vector which is named $\bm{x}$. 
\begin{equation}
\label{regd}
\min_{\bm{x}} \{{\lVert \bm{y}-\bm{Dx} \rVert}^2_2\}
\end{equation}
Since there is no space constraint, the solution can be reached rapidly in close- form as indicated in equation \eqref{resd}
\begin{equation}
\label{resd}
\bm{x} = (\bm{D}^T\bm{D})^{-1}\bm{D}^T\bm{y}
\end{equation}
The result vector consists of two part, the coefficients corresponding first class, and the ones corresponding second class. 
The sparseness will be involved from now, because this method uses all elements of the coefficient vector, but the final decision will be weighted. This means that the elements with larger corresponding Eigenvalues will have a greater role in decision making, and will lead to the sparseness properties.
The method is described in following equations in detail:
\begin{equation}
\begin{cases}
\sum_{i=1}^{n_1}\bm{x}_1(i)\lambda_1(i) > \sum_{i=1}^{n_2}\bm{x}_2(i)\lambda_2(i) \Rightarrow Class 1 \\
\sum_{i=1}^{n_1}\bm{x}_1(i)\lambda_1(i) \leq \sum_{i=1}^{n_2}\bm{x}_2(i)\lambda_2(i) \Rightarrow Class 2
\end{cases}
\end{equation}
In the above equations,$n_1$ is the number of the chosen bases for first class, and $\lambda$ is a vector containing the Eigenvalues from SVD of first class's dictionary.

Afterwards, it will be proved that weighting the coefficient for decision making is the same as training a new dictionary with weighted basis, but inversely proportional to the eigenvalues.

\section{Analytic Discussion}
Following equations indicate that using $\bm{D}_{new}$ instead of d is the same as weighting the result by eigenvalues. in below equations D is the dictionary matrix, $\lambda$ is a diagonal dictionary with the eigenvalues in its diagonal.
\begin{equation}
\bm{x}_{new} = (\bm{D}_{new}^T\bm{D}_{new})^{-1}\bm{D}_{new}^T\bm{y}, \bm{D}_{new} = \bm{D}\Sigma^{-1}
\end{equation} 

\begin{equation}
\bm{x}_{new} = ((\Sigma^{-1})^T\bm{D}^T\bm{D}\Sigma^{-1})^{-1}(\Sigma^{-1})^{T}\bm{D}^T
\end{equation}

\begin{equation}
\bm{x}_{new} = (\bm{D}^T\bm{D}\Sigma^{-1})^{-1}\Sigma^T(\Sigma^T)^{-1}\bm{D}^T
\end{equation}

\begin{equation}
\bm{x}_{new} = \Sigma(\bm{D}^T\bm{D})^{-1}\bm{D}^T
\end{equation}

\begin{equation}
\bm{x}_{new} = \Sigma\bm{x}
\end{equation}

%\begin{figure}[h]
%\includegraphics[width=.5\textwidth]{discus}
%\caption{Comparing Eigenvalue, and Eigenvectors before, and after dictionary learning}
%\end{figure}

\section{Experimental Results}
Using both experimental and theoretical data is necessary to investigate the performance a classification method. 
The WORM have been tested by BioMag 2012 competition data . This method have increased the percent of that international competition significantly.
Our results comparing to previous report from winner of BioMag 2012 international competition have been shown in Table 1. 
Until now, wide types of features have been used in classifying MEG signal in papers . Two new features have introduced in this paper that have shown better results than previous ones are differential spectrum and correlation between two adjacent channels in Fourier space. 
\begin{table}[H]
\caption{Testing WORM on experimental data}
\includegraphics[width=.5\textwidth]{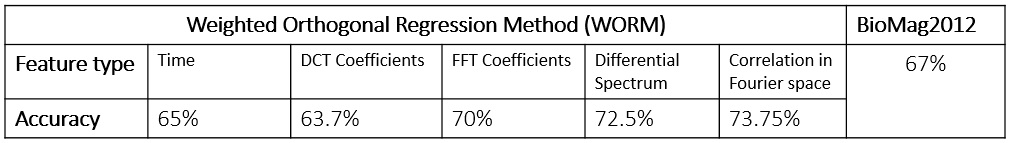}
\end{table}
To investigate the performance of WORM, three different types of theoretical data was made. Those data was deviated 200 dimensional points of 30 lines in 30 different direction using three different types of noise. The number of the training and testing data are respectively 200 and 1000. Selecting low number of training data is an advantage for that theoretical data since with such a small number of training data using a classification method that works based on PDF estimation will not be possible.
Three types of noise such as pepper and salt, Gaussian, and multiplicative noise have been tested. The performance of WORM in classifying those three types of data has been compared to other classifiers such as (K Nearest Neighbor)KNN, (Support Vector Machine)SVM, and (Orthogonal Matching Pursuit)OMP\cite{OMP}. 
As expected, the classification accuracy will reduced by reducing the signal to noise level which is shown in tables.

\begin{figure}[H]
\includegraphics[width=.5\textwidth]{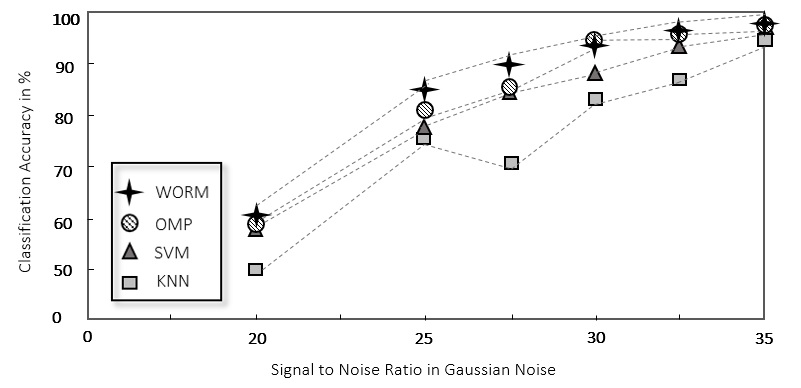}
\caption{Testing WORM accuracy on theoretical data with Gaussian noise}
\end{figure}

\begin{figure}[H]
\includegraphics[width=.5\textwidth]{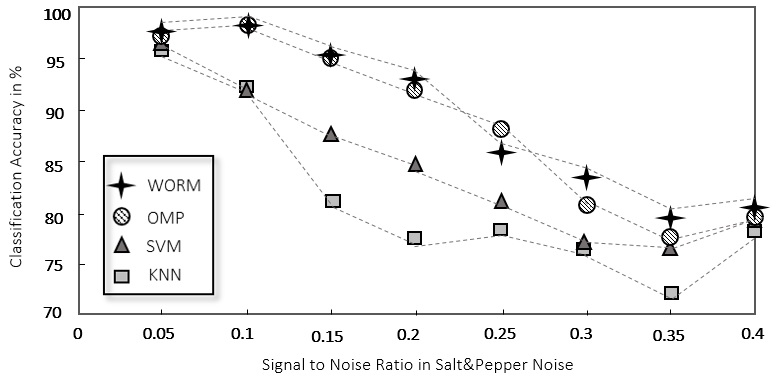}
\caption{Testing WORM accuracy on theoretical data with Pepper, and Salt noise }
\end{figure}

\begin{figure}[H]
\includegraphics[width=.5\textwidth]{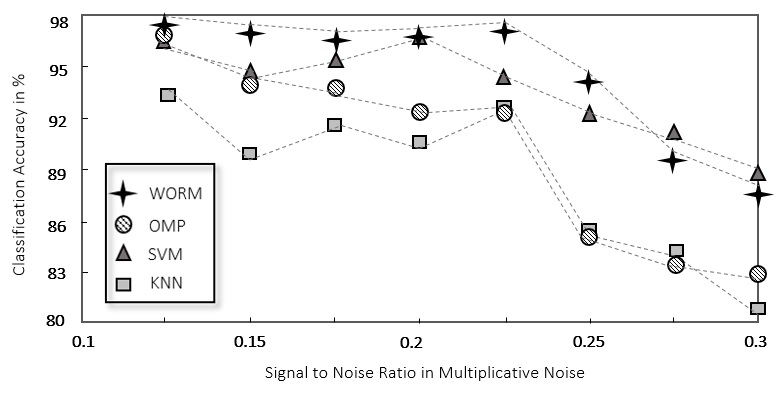}
\caption{Testing WORM accuracy on theoretical data with Multiplied noise}
\end{figure}

As it can be seen in above tables, WORM has shown a better performance especially in low signal to noise cases. The advantage of this method to OMP method which in some case have the same percent result is the higher speed of WORM because of applying sparseness after solving regression problem not before it.
\section{Conclusion}
In summary, the new method of classification have been introduced in this paper which is named WORM. The bases of this method is on regression problems, and applying the sparseness to them in a new way that increases the speed of processing as well as accuracy.  WORM have reached the better result comparing to other famous method in case of facing a classification problem with limited training set. Rather than being successful in testing on artificial data, this method has increased the best result of an international competition.

\end{document}